\documentclass[12pt]{article}
\usepackage{authblk}
\usepackage{graphicx}
\usepackage{dcolumn}
\usepackage{tensor}
\usepackage{bm}
\usepackage{floatrow}
\usepackage{color}
\usepackage{listings}
\usepackage{slashed}
\usepackage{amsmath}
\usepackage[usenames,dvipsnames]{xcolor}
\usepackage{amssymb}
\usepackage{amsfonts}
\usepackage{enumitem}
\usepackage{tensor}
\usepackage{mathtools}
\usepackage{slashed}
\usepackage{verbatim}
\usepackage{feynmf}
\usepackage{hyperref}
\usepackage{mathrsfs,hyperref,color,float}
\hypersetup{colorlinks=true, linkcolor=blue}
\usepackage{cancel}
\usepackage{subfig}
\numberwithin{equation}{section}

\newcommand{\be}{\begin{equation}}
\newcommand{\ee}{\end{equation}}

\begin{document}
\author[ ]{Sean A. Cantrell}
\affil[ ]{Excella Consulting AI Research}
\affil[ ]{\textit{sean.cantrell@excella.com}}
\title{The emergent algebraic structure of RNNs and embeddings in NLP}
\date{}
\maketitle

\begin{abstract}
We examine the algebraic and geometric properties of a uni-directional GRU and word embeddings trained end-to-end on a text classification task. A hyperparameter search over word embedding dimension, GRU hidden dimension, and a linear combination of the GRU outputs is performed.  We conclude that words naturally embed themselves in a Lie group and that RNNs form a nonlinear representation of the group. Appealing to these results, we propose a novel class of recurrent-like neural networks and a word embedding scheme.
\end{abstract}

\section{Introduction}
Tremendous advances in natural language processing (NLP) have been enabled by novel deep neural network architectures and word embeddings. Historically, convolutional neural network (CNN)\cite{lecun1998gradient,Zhang:charcnn} and recurrent neural network (RNN)\cite{elman1990finding,Yang:han} topologies have competed to provide state-of-the-art results for NLP tasks, ranging from text classification to reading comprehension.  CNNs identify and aggregate patterns with increasing feature sizes, reflecting our common practice of identifying patterns, literal or idiomatic, for understanding language; they are thus adept at tasks involving key phrase identification.  RNNs instead construct a representation of sentences by successively updating their understanding of the sentence as they read new words, appealing to the formally sequential and rule-based construction of language.  While both networks display great efficacy at certain tasks \cite{yin2017comparative}, RNNs tend to be the more versatile, have emerged as the clear victor in, e.g., language translation \cite{Britz:seq2seq, chiu2017monotonic, Vaswani:attn}, and are typically more capable of identifying important contextual points through attention mechanisms for, e.g., reading comprehension \cite{Cui:comprehension, Xiong:comprehension, Kumar:dmn, Xiong:dmn+}.  With an interest in NLP, we thus turn to RNNs.

RNNs nominally aim to solve a general problem involving sequential inputs. For various more specified tasks, specialized and constrained implementations tend to perform better \cite{Ruben:nested, Arjovsky:unitary, Wisdom:unitary, Vaswani:attn, Hyland:unitary, costa2017cortical, Jing:gorthornn, Kumar:dmn, Xiong:dmn+, Cui:comprehension, Xiong:comprehension}. Often, the improvement simply mitigates the exploding/vanishing gradient problem \cite{ Chung:gru, Hochreiter:lstm}, but, for many tasks, the improvement is more capable of generalizing the network's training for that task. Understanding better how and why certain networks excel at certain NLP tasks can lead to more performant networks, and networks that solve new problems.

Advances in word embeddings have furnished the remainder of recent progress in NLP \cite{Mokolov:word2vec, Pennington:glove, Nickel:poincare, Speer:conceptnet, Mikolov:pretrained, Wu:starspace}. Although it is possible to train word embeddings end-to-end with the rest of a network, this is often either prohibitive due to exploding/vanishing gradients for long corpora, or results in poor embeddings for rare words \cite{Bahdanau:e2e}.  Embeddings are thus typically constructed using powerful, but heuristically motivated, procedures to provide pre-trained vectors on top of which a network can be trained.  As with the RNNs themselves, understanding better how and why optimal embeddings are constructed in, e.g., end-to-end training can provide the necessary insight to forge better embedding algorithms that can be deployed pre-network training.

Beyond improving technologies and ensuring deep learning advances at a breakneck pace, gaining a better understanding of how these systems function is crucial for allaying public concerns surrounding the often inscrutable nature of deep neural networks.  This is particularly important for RNNs, since nothing comparable to DeepDream or Lucid exists for them \cite{olah2018the}.

To these ends, the goal of this work is two fold. First, we wish to understand any emergent algebraic structure RNNs and word embeddings, trained end-to-end, may exhibit.  Many algebraic structures are well understood, so any hints of structure would provide us with new perspectives from which and tools with which deep learning can be approached.  Second, we wish to propose novel networks and word embedding schemes by appealing to any emergent structure, should it appear.

The paper is structured as follows.  Methods and experimental results comprise the bulk of the paper, so, for faster reference, \S \ref{sec:summary} provides a convenient summary and intrepretation of the results, and outlines a new class of neural network and new word embedding scheme leveraging the results. \S \ref{sec:motiv_and_setup} motivates the investigation into algebraic structures and explains the experimental setup.  \S \ref{sec:results} Discusses the findings from each of the experiments. \S \ref{sec:discussion} interprets the results, and motivates the proposed network class and word embeddings. \S \ref{sec:closing} provides closing remarks and discusses followup work, and \S \ref{sec:acknowledgements} gives acknowledgments.

To make a matter of notation clear going forward, we begin by referring to the space of words as $W$, and transition to $G$ after analyzing the results in order to be consistent with notation in the literature on algebraic spaces.

\section{Summary of results} \label{sec:summary}
We embedded words as vectors and used a uni-directional GRU connected to a dense layer to classify the account from which tweets may have originated.  The embeddings and simple network were trained end-to-end to avoid imposing any artificial or heuristic constraints on the system.

There are two primary takeaways from the work presented herein:
\begin{itemize}
\item Words naturally embed as elements in a Lie group, $G$, and end-to-end word vectors may be related to the generating Lie algebra.

\item RNNs learn to parallel transport nonlinear representations of $G$ either on the Lie group itself, or on a principal $G$-bundle.
\end{itemize}

The first point follows since 1) words are embedded in a continuous space; 2) an identity word exists that causes the RNN to act trivially on a hidden state; 3) word inverses exist that cause the RNN to undo its action on a hidden state; 4) the successive action of the RNN using two words is equivalent to the action of the RNN with a single third word, implying the multiplicative closure of words; and 5) words are not manifestly closed under any other binary action.

The second point follows given that words embed on a manifold, sentences traces out paths on the manifold, and the difference equation the RNN solves bears a striking resemble to the first order equation for parallel transport,
\begin{align}
(h_{t+1} - h_t) + \gamma_{w_t} h_t =& 0 \\
\gamma_{w_t} \equiv& 1 - R_{w_t},
\end{align}
where $h_t$ is the $t$-th hidden state encountered when reading over a sentence and $R_{w_t}$ is the RNN conditioned by the $t$-th word, $w_t$, acting on the hidden state.  Since sentences trace out a path on the word manifold, and parallel transport operators for representations of the word manifold take values in the group, the RNN must parallel transport hidden states either on the group itself or on a base space, $M$, equipped with some word field, $w:M\to G$, that connects the path in the base space to the path on the word manifold.

Leveraging these results, we propose two new technologies.

First, we propose a class of recurrent-like neural networks for NLP tasks that satisfy the differential equation
\be
D^n_t h(t) = 0,
\ee
where
\begin{align}
D_t = \partial_t + v^\mu(t) \Gamma_\mu,
\end{align}
and where $\Gamma$ and $v$ are learned functions.  $n=1$ corresponds to traditional RNNs, with $v^\mu \Gamma_\mu \propto \gamma$. For $n>1$, this takes the form of RNN cells with either nested internal memories or dependencies that extend temporally beyond the immediately previous hidden state.  In particular, using $n=2$ for sentence generation is the topic of a manuscript presently in preparation.

Second, we propose embedding schemes that explicitly embed words as elements of a Lie group. In practice, these embedding schemes would involve representing words as constrained matrices, and optimizing the elements, subject to the constraints, according to a loss function constructed from invariants of the matrices, and then applying the matrix log to obtain Lie vectors. A prototypical implementation, in which the words are assumed to be in the fundamental representation of the special orthogonal group, $SO(N)$, and are conditioned on losses sensitive to the relative actions of words, is the subject of another manuscript presently in preparation.

The proposals are only briefly discussed herein, as they are the focus of followup work; the focus of the present work is on the experimental evidence for the emergent algebraic structure of RNNs and embeddings in NLP.

\section{Motivation and experimental setup} \label{sec:motiv_and_setup}
\subsection{Intuition and motivation} \label{sec:motivation}
We provide two points to motivate examining the potential algebraic properties of RNNs and their space of inputs in the context of NLP.

First, a RNN provides a function, $R$, that successively updates a hidden memory vector, $h \in H$, characterizing the information contained in a sequence of input vectors, $\left\{w_1, w_2, \dots \right\} \in W$, as it reads over elements of the sequence. Explicitly, $R : W \times H \to H$.  At face value, $R$ takes the same form as a (nonlinear) representation of some general algebraic structure, $W$, with at least a binary action, $\cdot : W \times W \to W$, on the vector space $H$.  While demanding much structure on $W$ generally places a strong constraint on the network's behavior, it would be fortuitous for such structure to emerge. Generally, constrained systems still capable of performing a required task will perform the task better, or, at least, generalize more reliably \cite{pascanu2013number, montufar2014number, livni2014computational, telgarsky2016benefits, poggio2017and, kawaguchi2017generalization}. To this end, the suggestive form RNNs assume invites further examination to determine if there exist any reasonable constraints that may be placed on the network.  To highlight the suggestiveness of this form in what follows, we represent the $W$ argument of $R$ as a subscript and the $H$ argument by treating $R$ as a left action on $H$, adopting the notation $R(w, h) \to R_w h$. Since, in this paper, we consider RNNs vis-\`{a}-vis NLP, we take $W$ as the (continuous) set of words\footnote{Traditionally, words are treated as categorical objects, and embedding them in a continuous (vector) space for computational purposes is largely a convenience; however, we relax this categorical perspective, and treat unused word vectors as acceptable objects as far as the algebraic structure is concerned, even if they are not actively employed in language.}.

Second, in the massive exploration of hyperparameters presented in \cite{Britz:seq2seq}, it was noted that, for a given word embedding dimension, the network's performance on a seq2seq task was largely insensitive to the hidden dimension of the RNN above a threshold ($\sim$128).  The dimension of admissible representations of a given algebraic structure is generally discrete and spaced out.  Interpreting neurons as basis functions and the output of layers as elements of the span of the functions \cite{Raghu:svcca, csaji2001approximation, hornik1991approximation}, we would expect a network's performance to improve until an admissible dimension for the representation is found, after which the addition of hidden neurons would simply contribute to better learning the components of the proper representation by appearing in linear combinations with other neurons, and contribute minimally to improving the overall performance.  In their hyperparameter search, a marginal improvement was found at a hidden dimension of 2024, suggesting a potentially better representation may have been found.

These motivating factors may hint at an underlying algebraic structure in language, at least when using RNNs, but they raise the question: what structures are worth investigating?

Groups present themselves as a candidate for consideration since they naturally appear in a variety of applications.  Unitary weight matrices have already enjoyed much success in mitigating the exploding/vanishing gradients problem \cite{Arjovsky:unitary,Wisdom:unitary}, and RNNs even further constrained to act explicitly as nonlinear representations of unitary groups offer competitive results \cite{Hyland:unitary}.  Moreover, intuitively, RNNs in NLP could plausibly behave as a group since: 1) the RNN must learn to ignore padding words used to square batches of training data, indicating an identity element of $W$ must exist; 2) the existence of contractions, portmanteaus, and the Germanic tradition of representing sentences as singular words suggest $W$ might be closed; and 3) the ability to backtrack and undo statements suggests language may admit natural inverses - that is, active, controlled ``forgetting" in language may be tied to inversion. Indeed, groups seem reasonably promising.

It is also possible portmanteaus only make sense for a finite subset of pairs of words, so $W$ may take on the structure of a groupoid instead; moreover, it is possible, at least in classification tasks, that information is lost through successive applications of $R$, suggesting an inverse may not actually exist, leaving $W$ as either a monoid or category.  $W$ may also actually admit \textit{additional} structure, or an additional binary operation, rendering it a ring or algebra.

To determine what, if any, algebraic structure $W$ possesses, we tested if the following axiomatic properties of faithful representations of $W$ hold:
\begin{enumerate}
\item (Identity) $\exists I \in W$ such that $\forall h \in H$, $R_I h  = h$
\item (Closure under multiplication) $\forall w_1, w_2 \in W$, $\exists w_3 \in W$ such that $\forall h \in H$, $R_{w_2} R_{w_1} h = R_{w_3} h$
\item (Inverse) $\forall w \in W$, $\exists w^{-1} \in W$ such that $\forall h \in H$, $R_{w^{-1}} R_w h = R_w R_{w^{-1}} h = h$
\item (Closure under Lie bracket) $\forall w_1, w_2 \in W$, $\exists w_3 \in W$ such that $\forall h \in H$, $R_{w_2} R_{w_1} h - R_{w_1} R_{w_2} h = R_{w_3} h$
\end{enumerate}
Closure under Lie bracket simultaneously checks for ring and Lie algebra structures.

Whatever structure, if any, $W$ possesses, it must additionally be continuous since words are typically embedded in continuous spaces.  This implies Lie groups (manifolds), Lie semigroups with an identity (also manifolds), and Lie algebras (vector spaces with a Lie bracket) are all plausible algebraic candidates.

\subsection{Data and methods}
\begin{figure}[H] 
\begin{center}
\includegraphics[scale=0.7]{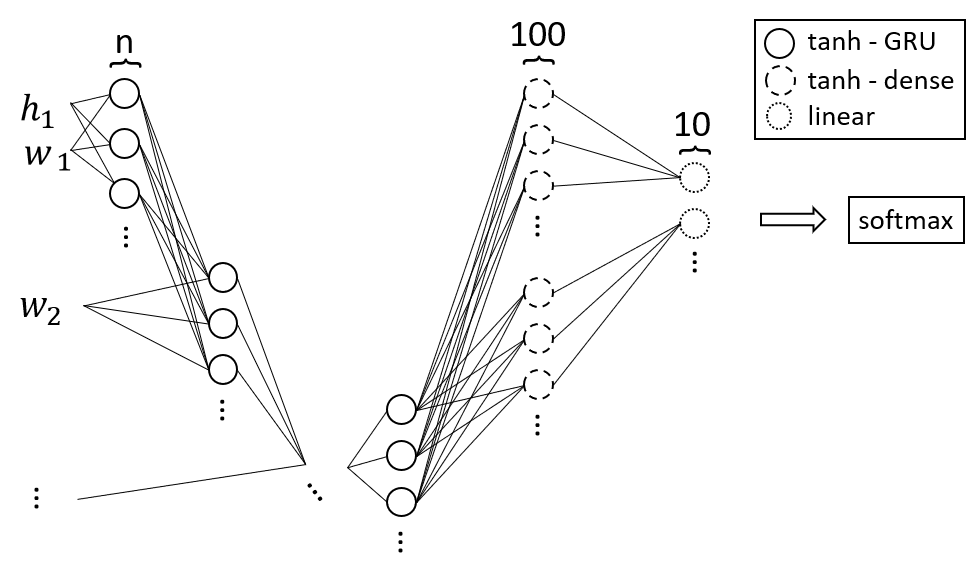} 
\caption{\footnotesize The simple network trained as a classifier: GRU$\to$Dense$\to$Linear$\to$Softmax.  There are 10 nonlinear neurons dedicated to each of the final 10 energies that are combined through a linear layer before softmaxing.  This is to capitalize on the universal approximation theorem's implication that neurons serve as basis functions - i.e. each energy function is determined by 10 basis functions. The hidden dimension $n$ of the GRU, and the word embedding dimension, are hyperparameters that are scanned over.} \label{fig:network}
\end{center}
\end{figure}

We trained word embeddings and a uni-directional GRU connected to a dense layer end-to-end for text classification on a set of scraped tweets using cross-entropy as the loss function. End-to-end training was selected to impose as few heuristic constraints on the system as possible. Each tweet was tokenized using NLTK TweetTokenizer and classified as one of 10 potential accounts from which it may have originated.  The accounts were chosen based on the distinct topics each is known to typically tweet about. Tokens that occurred fewer than 5 times were disregarded in the model. The model was trained on 22106 tweets over 10 epochs, while 5526 were reserved for validation and testing sets (2763 each). The network demonstrated an insensitivity to the initialization of the hidden state, so, for algebraic considerations, $(\frac{1}{\sqrt{n}}, \frac{1}{\sqrt{n}}, \dots )$ was chosen for hidden dimension of $n$.  A graph of the network is shown in Fig.(\ref{fig:network}).

Algebraic structures typically exhibit some relationship between the dimension of the structure and the dimension of admissible representations, so exploring the embedding and hidden dimensions for which certain algebraic properties hold is of interest. Additionally, beyond the present interest in algebraic properties, the network's insensitivity to the hidden dimension invites an investigation into its sensitivity to the word embedding dimension. To address both points of interest, we extend the hyperparameter search of \cite{Britz:seq2seq}, and perform a comparative search over embedding dimensions and hidden dimensions to determine the impact of each on the network's performance and algebraic properties.  Each dimension in the hyperparameter pair, $(m,n) = (\text{embedding dim}, \text{hidden dim})$, runs from 20 to 280 by increments of 20.

After training the network for each hyperparameter pair, the GRU model parameters and embedding matrix were frozen to begin testing for emergent algebraic structure. To satisfy the common ``$\forall h\in H$" requirement stated in \S \ref{sec:motivation}, real hidden states encountered in the testing data were saved to be randomly sampled when testing the actions of the GRU on states. 7 tests were conducted for each hyperparameter pair with randomly selected states:
\begin{enumerate}
\item Identity (``arbitrary identity")
\item Inverse of all words in corpus (``arbitrary inverse")
\item Closure under multiplication of arbitrary pairs of words in total corpus (``arbitrary closure")
\item Closure under commutation of arbitrary pairs of words in total corpus (``arbitrary commutativity")
\item Closure under multiplication of random pairs of words from within each tweet (``intra-sentence closure")
\item Closure of composition of long sequences of words in each tweet (``composite closure")
\item Inverse of composition of long sequences of words in each tweet (``composite inverse")
\end{enumerate}
Tests 6 and 7 were performed since, if closure is upheld, the composition of multiple words must also be upheld. These tests were done to ensure mathematical consistency.

To test for the existence of ``words" that satisfy these conditions, vectors were searched for that, when inserted into the GRU, minimized the ratio of the Euclidean norms of the difference between the ``searched" hidden vector and the correct hidden vector.  For concreteness, the loss function for each algebraic property from \S \ref{sec:motivation} were defined as follows:
\begin{enumerate}
\begin{subequations} \label{eqn:losses}
\item (Identity)
\begin{align}
\mathcal{L}_\text{id} =& \frac{|R_x h - h|}{|h|} \label{eqn:id_loss}
\end{align}

 \item (Closure under multiplication)
\begin{align}
\mathcal{L}_\text{closure} =& \frac{|R_{x} h - R_{w_2} R_{w_1} h|}{|R_{w_2} R_{w_1} h|}. \label{eqn:clo_loss}
\end{align}

\item (Inverse)
\begin{align}
\mathcal{L}_\text{inverse} =& \frac{|R_{x} R_w h - h|}{|h|} \label{eqn:inv_loss}
\end{align}

\item (Closure under Lie bracket)
\begin{align}
\mathcal{L}_\text{com} =& \frac{|R_x h - (R_{w_2} R_{w_1} h - R_{w_1} R_{w_2} h)|}{\max\left(|(R_{w_2} R_{w_1} h - R_{w_1} R_{w_2} h)|,10^{-6}\right)} \label{eqn:com_loss}
\end{align}
\end{subequations}
\end{enumerate}
where $w_i$ are random, learned word vectors, $h$ is a hidden state, and $x$ is the model parameter trained to minimize the loss.  We refer to Eqs.(\ref{eqn:losses}) as the ``axiomatic losses."  It is worth noting that the non-zero hidden state initialization was chosen to prevent the denominators from vanishing when the initial state is selected as a candidate $h$ in Eqs.(\ref{eqn:id_loss})\&(\ref{eqn:inv_loss}).  The reported losses below are the average across all $w$'s and $h$'s that were examined. Optimization over the losses in Eqs.(\ref{eqn:losses}) was performed over 5000 epochs. For the associated condition to be satisfied, there must exist a word vector $x$ that sufficiently minimizes the axiomatic losses.

If it is indeed the case that the GRU attempts to learn a representation of an algebraic structure and each neuron serves as a basis function, it is not necessary that each neuron individually satisfies the above constraints.  For clarity, recall the second motivating point that the addition of neurons, once a representation is found, simply contributes to learning the representation better. Instead, only a linear combination of the neurons must.  We consider this possibility for the most task-performant hyperparameter pair, and two other capricious pairs.  The target dimension of the linear combination, $p$, which we refer to as the ``latent dimension," could generally be smaller than the hidden dimension, $n$.  To compute the linear combination of the neurons, the outputs of the GRU were right-multiplied by a $n \times p$ matrix, $P$\footnote{There may be concern over the differing treatment of the output of the GRU and the input hidden state, since the former is being projected into a lower dimension while the latter is not. If the linear combination matrix were trained in parallel to the GRU itself, there would be a degeneracy in the product between it and the GRU weight matrices in successive updates, and it and the dense layer weight matrices after the final update, such that the effect of the linear combination would be absorbed by the weight matrices.  To this end, since we are considering linear combinations after freezing the GRU weight matrices, it is unnecessary to consider the role the linear combination matrix would play on the input hidden states, and necessary for only the output of the GRU itself.}:
\begin{subequations}
\begin{align} \label{eqn:lin_comb}
h \to& h P \\
R_w h \to& (R_w h) P \\
R_{w_2} R_{w_1} h \to& (R_{w_2} R_{w_1} h) P
\end{align}
\end{subequations}
Since the linear combination is not \`{a} priori known, $P$ is treated as a model parameter.

The minimization task previously described was repeated with this combinatorial modification while scanning over latent dimensions, $p \in [20,n-20]$, in steps of 20.  The test was performed 10 times and the reported results averaged for each value of $p$ to reduce fluctuations in the loss from differing local minima.  $P$ was trained to optimize various combinations of the algebraic axioms, the results of which were largely found to be redundant.  In \S \ref{sec:results}, we address the case in which $P$ was only trained to assist in optimizing a single condition, and frozen in other axiomatic tests; the commutative closure condition, however, was given a separate linear combination matrix for reasons that will be discussed later.

Finally, the geometric structure of the resulting word vectors was explored, naively using the Euclidean metric.  Sentences trace out (discrete) paths in the word embedding space, so it was natural to consider relationships between both word vectors and vectors ``tangent" to the sentences' paths. Explicitly, the angles and distances between
\begin{enumerate}
\item random pairs of words

\item all words and the global average word vector

\item random pairs of co-occurring words

\item all words with a co-occurring word vector average

\item adjacent tangent vectors

\item tangent vectors with a co-occurring tangent vector average
\end{enumerate}
were computed to determine how word vectors are geometrically distributed.  Intuitively, similar words are expected to affect hidden states similarly.  To test this, and to gain insight into possible algebraic interpretations of word embeddings, the ratio of the Euclidean norm of the difference between hidden states produced by acting on a hidden state with two different words to the Euclidean norm of the original hidden state was computed as a function of the popular cosine similarity metric and distance between embeddings.  This fractional difference, cosine similarity, and word distance were computed as,
\begin{align}
\mathfrak{E} =& \frac{|R_{w_1} h - R_{w_2} h|}{|h|} \label{eqn:frac_dist} \\
\cos(\theta_w) =& w_1^\alpha w_2^\alpha, \label{eqn:cos_sim}\\
|\Delta w| =& ||w_1 - w_2||_2 \label{eqn:dist_sim},
\end{align}
where Einstein summation is applied to the (contravariant) vector indices.

High-level descriptions of the methods will be briefly revisited in each subsection of \S \ref{sec:results} so that they are more self-contained and pedagogical.

\section{Results} \label{sec:results}
\subsection{Hyperparameters and model accuracy}
\begin{figure}[H] 
\begin{center}
\includegraphics[scale=0.5]{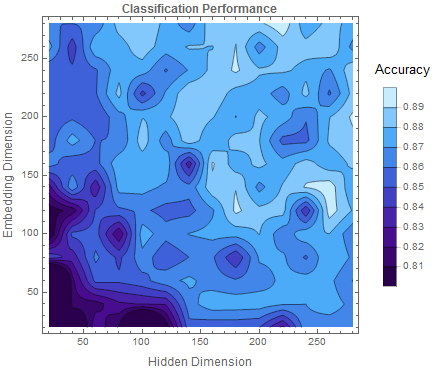}
\caption{\footnotesize The range of the model accuracy is $[50.1\%, 89.7\%]$.} \label{fig:class_acc}
\end{center}
\end{figure}

We performed hyperparameter tuning over the word embedding dimension and the GRU hidden dimension to optimize the classifier's accuracy.  Each dimension ran from 20 to 280 in increments of 20.  A contour plot of the hyperparameter search is shown in Fig.(\ref{fig:class_acc}).

For comparison, using pretrained, 50 dimensional GloVe vectors with this network architecture typically yielded accuracies on the order of $\mathcal{O}(50\%)$ on this data set, even for more performant hidden dimensions. Thus, training the embeddings end-to-end is clearly advantageous for short text classification.  It is worth noting that training them end-to-end is viable primarily because of the short length of tweets; for longer documents, exploding/vanishing gradients typically prohibits such training.

The average Fisher information of each hyperparameter dimension over the searched region was computed to determine the relative sensitivities of the model to the hyperparameters. The Fisher information for the hidden dimension was $4.63 \times 10^{-6}$; the Fisher information for the embedding dimension was $2.62 \times 10^{-6}$.  Evidently, by this metric, the model was, on average in this region of parameter space, 1.76 times more sensitive to the hidden dimension than the embedding dimension. Nevertheless, a larger word embedding dimension was critical for the network to realize its full potential.

The model performance generally behaved as expected across the hyperparameter search.  Indeed, higher embedding and hidden dimensions tended to yield better results. Given time and resource constraints, the results are not averaged over many search attempts.  Consequently, it is unclear if the pockets of entropy are indicative of anything deeper, or merely incidental fluctuations.  It would be worthwhile to revisit this search in future work.

\subsection{Algebraic properties}
Seven tests were conducted for each hyperparameter pair to explore any emergent algebraic structure the GRU and word embeddings may exhibit.  Specifically, the tests searched for 1) the existence of an identity element, 2) existence of an inverse word for each word, 3) multiplicative closure for arbitrary pairs of words, 4) commutative closure for arbitrary pairs of words, 5) multiplicative closure of pairs of words that co-occur within a tweet, 6) multiplicative closure of all sequences of words that appear in tweets, and 7) the existence of an inverse for all sequences of words that appear in tweets.  The tests optimized the axiomatic losses defined in Eqs.(\ref{eqn:losses}).

In what follows, we have chosen $\mathcal{L}<0.01$ (or, $1\%$ error) as the criterion by which we declare a condition ``satisfied."

\begin{figure}[H]
\begin{center}
\begin{tabular}{cc}
\sidesubfloat[]{\includegraphics[scale=0.5]{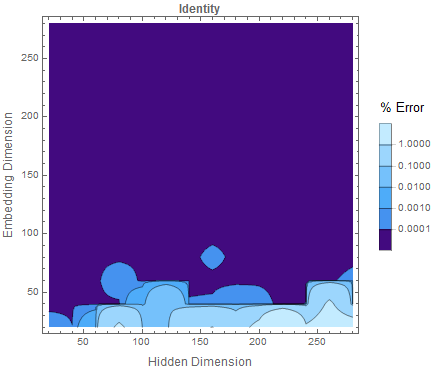}}&
\sidesubfloat[]{\includegraphics[scale=0.5]{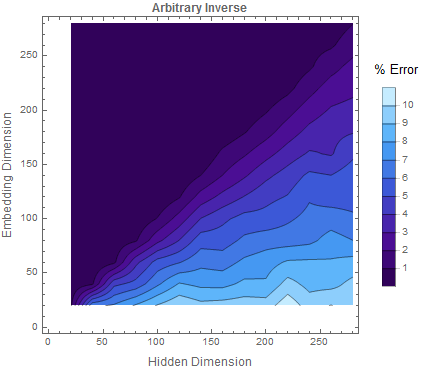}}\\
\sidesubfloat[]{\includegraphics[scale=0.5]{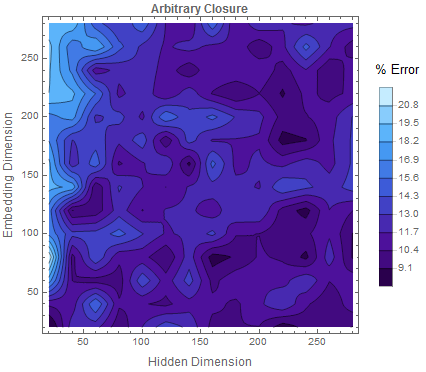}}&
\sidesubfloat[]{\includegraphics[scale=0.5]{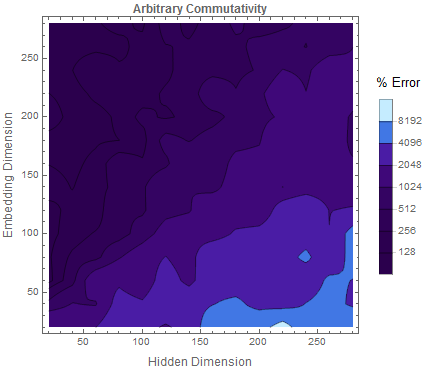}}
\end{tabular}
\caption{\footnotesize The \% axiomatic error as a function of the word embedding and GRU hidden dimensions. (a) The existence of an identity element for multiple hidden states. Note the log scale. (b) The existence of an inverse word for every word acting on random hidden states. Linear scale. (c) The existence of a third, `effective' word performing the action of two randomly chosen words in succession, acting on random states. Linear scale. (d) The existence of a third word performing the action of the commutation of two randomly chosen words, acting on random states. Nonlinear scale.} \label{fig:arb_elems}
\end{center}
\end{figure}

\begin{figure}[H] 
\begin{center}
\begin{tabular}{cc}
\sidesubfloat[]{\includegraphics[scale=0.5]{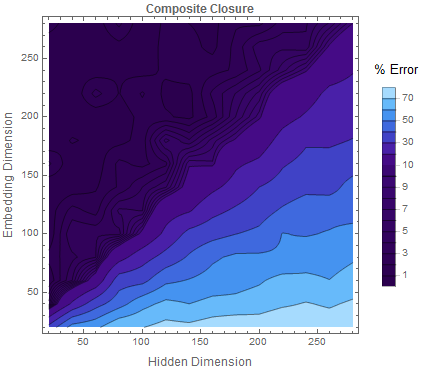}}&

\sidesubfloat[]{\includegraphics[scale=0.5]{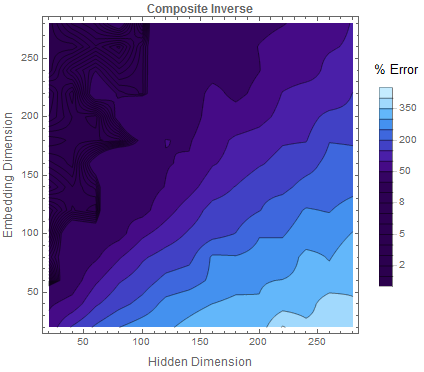}}\\
\sidesubfloat[]{\includegraphics[scale=0.5]{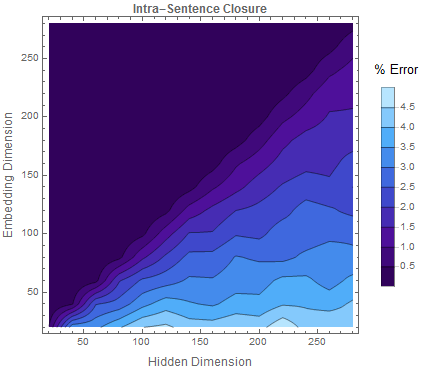}}&
\sidesubfloat[]{\includegraphics[scale=0.5]{./graphics/arb_inverse.png}}
\end{tabular}
\caption{\footnotesize The \% axiomatic error as a function of the word embedding and GRU hidden dimensions. (a)  The existence of a third word performing the action of all, ordered words comprising a tweet, acting on the initial state. Linear scale. (b) The existence of a word that reverses the action of the ordered words comprising a tweet that acted on the initial state. Nonlinear scale. (c) The existence of a third word performing the action of two random words co-occurring within a tweet, acting on random states. Linear scale. (d)  The existence of an inverse word for every word acting on random hidden states. This is the same as in Fig.(\ref{fig:arb_elems}), and is simply provided for side-by-side comparison.} \label{fig:sent_elems}
\end{center}
\end{figure}

The tests can be broken roughly into two classes: 1) arbitrary solitary words and pairs of words, and 2) pairs and sequences of words co-occurring within a tweet.  The results for class 1 are shown in Fig.(\ref{fig:arb_elems}); the results for class 2 are shown in Fig.(\ref{fig:sent_elems}).

The identity condition was clearly satisfied for virtually all embedding and hidden dimensions, with possible exceptions for small embedding dimensions and large hidden dimensions.  Although we did not explicitly check, it is likely that even the possible exceptions would be viable in the linear combination search.

Arbitrary pairs of words were evidently not closed under multiplication without performing a linear combination search, with a minimum error of $7.51\%$ across all dimensions.  Moreover, the large entropy across the search does not suggest any fundamentally interesting or notable behavior, or any connections between the embedding dimension, hidden dimension, and closure property.

Arbitrary pairs of words were very badly not closed under commutation, and it is unfathomable that even a linear combination search could rescue the property. One might consider the possibility that specific pairs of words might have still closed under commutation, and that the exceptional error was due to a handful of words that commute outright since this would push the loss up with a near-vanishing denominator.  As previously stated, the hidden states were not initialized to be zero states, and separate experiments confirm that the zero state was not in the orbit of any non-zero state, so there would have been no hope to negate the vanishing denominator. Thus, this concern is in principle possible.  However, explicitly removing examples with exploding denominators (norm$<10^{-3}$) from the loss when performing linear combination searches still resulted in unacceptable errors ($80\% +$), so this possibility is not actually realized.  We did not explicitly check for this closure in class 2 tests since class 2 is a subset of class 1, and such a flagrant violation of the condition would not be possible if successful closure in class 2 were averaged into class 1 results. Even though commutative closure is not satisfied, it is curious to note that the error exhibited a mostly well-behaved stratification.

The most interesting class 1\footnote{The arbitrary inverse is neither, strictly, class 1 nor class 2 since it does not involve pairings with any other words. We simply group it with class 1 to keep it distinct from the composite inverse experiment, which is decidedly class 2.} result was the arbitrary inverse. For embedding dimensions sufficiently large compared to the hidden dimension, inverses clearly existed even without a linear combination search.  Even more remarkable was the well-behaved stratification of the axiomatic error, implying a very clear relationship between the embedding dimension, hidden dimension, and emergent algebraic structure of the model.  It is not unreasonable to expect the inverse condition to be trivially satisfied in a linear combination search for a broad range of hyperparameter pairs.

The same behavior of the inverse property is immediately apparent in all class 2 results.  The stratification of the error was virtually identical, and all of the tested properties have acceptable errors for sufficiently large embedding dimensions for given hidden dimensions, even without a linear combination search.

\subsection{Linear combination search}
The optimal hyperparameter pair for this single pass of tuning was $(m,n)=(280,220)$, which resulted in a model accuracy of $89.7\%$.  This was not a statistically significant result since multiple searches were not averaged, so random variations in validation sets and optimization running to differing local minima may have lead to fluctuations in the test accuracies. However, the selection provided a reasonable injection point to investigate the algebraic properties of linear combinations of the output of the GRU's neurons.  For comparison, we also considered $(m,n)=(180,220)$ and $(m,n)=(100,180)$.

The tests were run with the linear combination matrix, $P$, trained to assist in optimizing the composite inverse. The learned $P$ was then applied to the output hidden states for the other properties except for commutative closure, which was given its own linear combination matrix to determine if any existed that would render it an emergent property.

The combination was trained to optimize a single condition because, if there exists an optimal linear combination for one condition, and there indeed exists an underlying algebraic structure incorporating other conditions, the linear combination would be optimal for all other conditions.

\begin{figure}[H] 
\begin{center}
\begin{tabular}{cc}
\sidesubfloat[]{\includegraphics[scale=0.5]{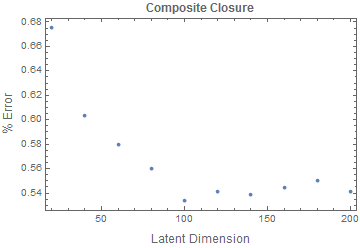}}&
\sidesubfloat[]{\includegraphics[scale=0.5]{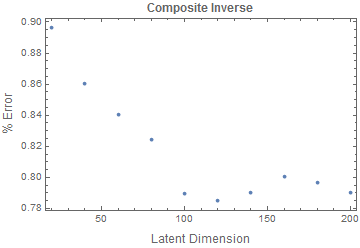}}\\
\sidesubfloat[]{\includegraphics[scale=0.5]{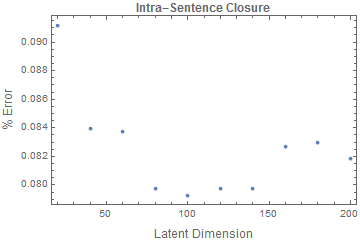}}&
\sidesubfloat[]{\includegraphics[scale=0.5]{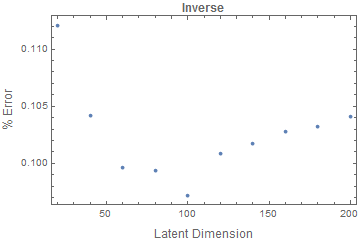}}
\end{tabular}
\caption{\footnotesize $(m,n) = (280,220)$. Graphs of \% axiomatic error for the satisfied conditions after a linear combination search. The graphs are ordered as they were in Fig.(\ref{fig:sent_elems})} \label{fig:lin_comb_grp_280_220}
\end{center}
\end{figure}

\begin{figure}[H] 
\begin{center}
\begin{tabular}{cc}
\sidesubfloat[]{\includegraphics[scale=0.5]{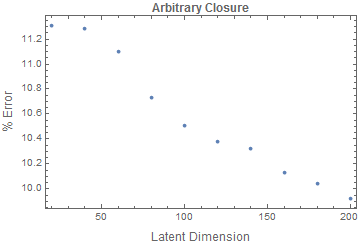}}&
\sidesubfloat[]{\includegraphics[scale=0.45]{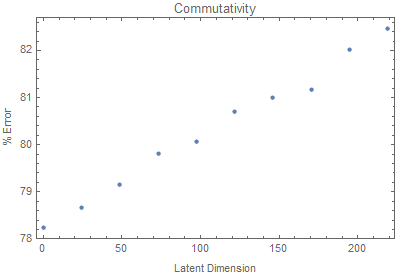}}
\end{tabular}
\caption{\footnotesize $(m,n) = (280,220)$. Graphs of \% axiomatic error for the unsatisfied conditions after a linear combination search.} \label{fig:lin_comb_bad_280_220}
\end{center}
\end{figure}

Initial results for the $(m,n) = (280,220)$ search is shown in Figs.(\ref{fig:lin_comb_grp_280_220})\&(\ref{fig:lin_comb_bad_280_220}).  Well-optimized properties are shown in Fig.(\ref{fig:lin_comb_grp_280_220}), while the expected poorly-optimized properties are shown in Fig.(\ref{fig:lin_comb_bad_280_220}).

The four conditions examined in Fig.(\ref{fig:lin_comb_grp_280_220}) are clearly satisfied for all latent dimensions. They all also reach a minimum error in the same region.  Composite closure, intra-sentence closure, and arbitrary inverse are all optimized for $p\approx 100$; composite inverse is optimized for $p\approx 120$, though the variation in the range $[100,140]$ is small ($\sim 0.8\%$ variation around the mean, or an absolute variation of $\sim 0.006\%$ in the error).  

Arbitrary multiplicative closure and commutative closure are highly anti-correlated, and both conditions are badly violated.  It is worth noting that the results in Fig.(\ref{fig:lin_comb_bad_280_220})(b) did not remove commutative pairs of words from the error, and yet the scale of the error in the linear combination search is virtually identical to what was separately observed with the commutative pairs removed. They both also exhibit a monotonic dependence on the latent dimension.  Despite their violation, this dependence is well-behaved, and potentially indicative of some other structure.

\begin{figure}[H] 
\begin{center}
\begin{tabular}{cc}
\sidesubfloat[]{\includegraphics[scale=0.5]{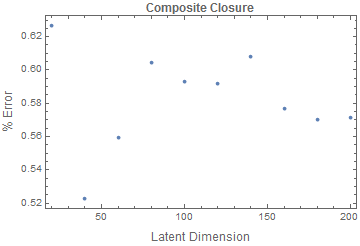}}&
\sidesubfloat[]{\includegraphics[scale=0.5]{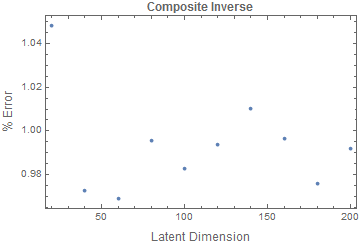}}\\
\sidesubfloat[]{\includegraphics[scale=0.5]{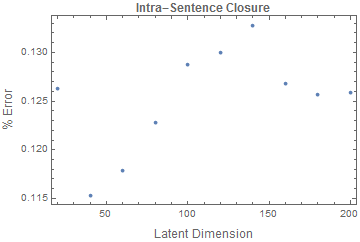}}&
\sidesubfloat[]{\includegraphics[scale=0.5]{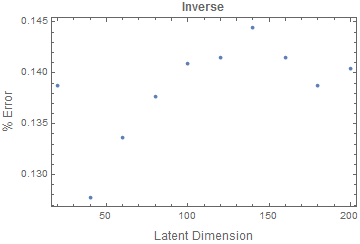}}
\end{tabular}
\caption{\footnotesize $(m,n) = (280,220)$, retrained. Graphs of \% axiomatic error for the satisfied conditions after a linear combination search. The graphs are ordered as they were in Fig.(\ref{fig:sent_elems})} \label{fig:lin_comb_grp_280_220_2}
\end{center}
\end{figure}

\begin{figure}[H] 
\begin{center}
\begin{tabular}{cc}
\sidesubfloat[]{\includegraphics[scale=0.5]{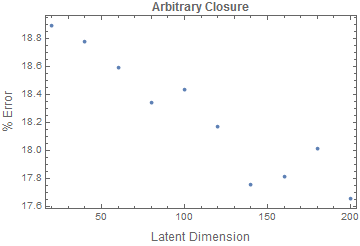}}&
\sidesubfloat[]{\includegraphics[scale=0.5]{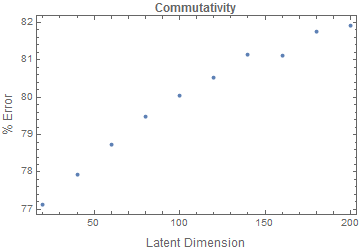}}
\end{tabular}
\caption{\footnotesize $(m,n) = (280,220)$, retrained. Graphs of \% axiomatic error for the unsatisfied conditions after a linear combination search.} \label{fig:lin_comb_bad_280_220_2}
\end{center}
\end{figure}

Before discussing the linear combination searches for the other selected hyperparameter pairs, it is worthwhile noting that retraining the network and performing the linear combination search again can yield differing results.  Figs.(\ref{fig:lin_comb_grp_280_220_2})\&(\ref{fig:lin_comb_bad_280_220_2}) show the linear combination results after retraining the model for the same hyperparameter pair, with a different network performance of $87.3\%$.

Qualitatively, the results are mostly the same: there is a common minimizing region of $p$, and conditions are satisfied, at least in the common minimal region.  However, the minimizing region starkly shifted down, and became sharper for composite closure, intra-sentence closure, and arbitrary inverse.  

Once more, the results are mostly the same.  Arbitrary closure error drastically increased, but both are still highly anti-correlated, and mostly monotonic, despite the erratic fluctuations in the arbitrary closure error.

\begin{figure}[H] 
\begin{center}
\begin{tabular}{cc}
\sidesubfloat[]{\includegraphics[scale=0.5]{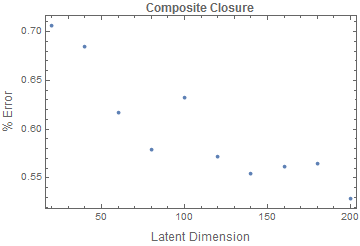}}&
\sidesubfloat[]{\includegraphics[scale=0.5]{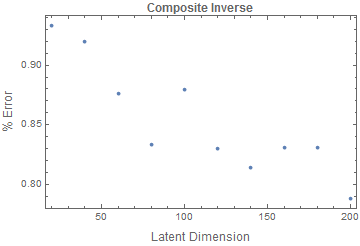}}\\
\sidesubfloat[]{\includegraphics[scale=0.5]{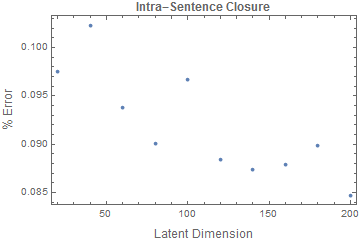}}&
\sidesubfloat[]{\includegraphics[scale=0.5]{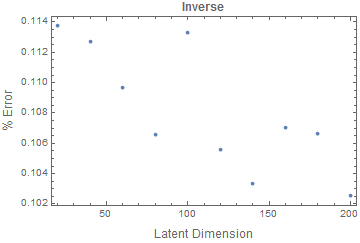}}
\end{tabular}
\caption{\footnotesize $(m,n) = (180,220)$. Graphs of \% axiomatic error for the satisfied conditions after a linear combination search. The graphs are ordered as they were in Fig.(\ref{fig:sent_elems})} \label{fig:lin_comb_grp_180_220}
\end{center}
\end{figure}

\begin{figure}[H] 
\begin{center}
\begin{tabular}{cc}
\sidesubfloat[]{\includegraphics[scale=0.5]{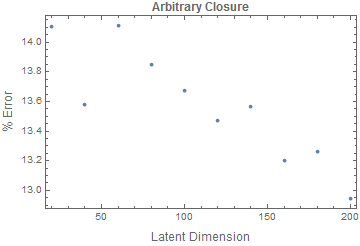}}&
\sidesubfloat[]{\includegraphics[scale=0.5]{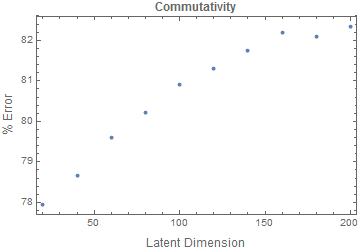}}
\end{tabular}
\caption{\footnotesize $(m,n) = (180,220)$. Graphs of \% axiomatic error for the unsatisfied conditions after a linear combination search.} \label{fig:lin_comb_bad_180_220}
\end{center}
\end{figure}

Figs.(\ref{fig:lin_comb_grp_180_220})\&(\ref{fig:lin_comb_bad_180_220}) show the linear combination search for $(m,n) = (180,220)$.  The model was retrained, and achieved $90.1\%$ for the displayed results.

Interestingly, the optimal latent dimension occurs significantly higher than for the other reported hyperparameter pairs.  This result, however, is not true for all retrainings at this $(m,n)$ pair.

The entropy in the arbitrary closure loss increased, and the commutative closure loss seemed to asymptote at higher latent dimension.

\begin{figure}[H] 
\begin{center}
\begin{tabular}{cc}
\sidesubfloat[]{\includegraphics[scale=0.5]{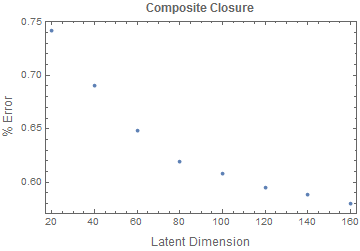}}&
\sidesubfloat[]{\includegraphics[scale=0.5]{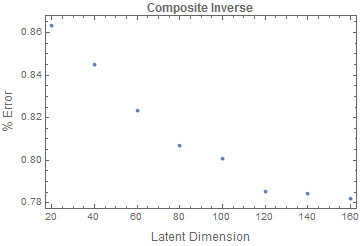}}\\
\sidesubfloat[]{\includegraphics[scale=0.5]{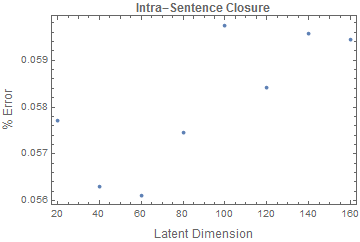}}&
\sidesubfloat[]{\includegraphics[scale=0.5]{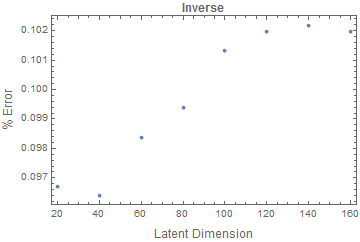}}
\end{tabular}
\caption{\footnotesize $(m,n) = (100,180)$. Graphs of \% axiomatic error for the satisfied conditions after a linear combination search. The graphs are ordered as they were in Fig.(\ref{fig:sent_elems})} \label{fig:lin_comb_grp_100_180}
\end{center}
\end{figure}

\begin{figure}[H] 
\begin{center}
\begin{tabular}{cc}
\sidesubfloat[]{\includegraphics[scale=0.5]{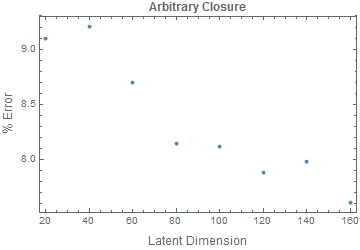}}&
\sidesubfloat[]{\includegraphics[scale=0.5]{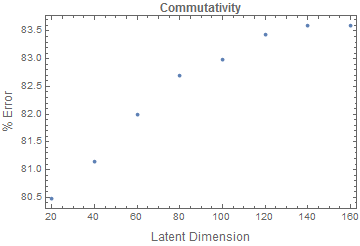}}
\end{tabular}
\caption{\footnotesize $(m,n) = (100,180)$. Graphs of \% axiomatic error for the unsatisfied conditions after a linear combination search.} \label{fig:lin_comb_bad_100_180}
\end{center}
\end{figure}

Figs.(\ref{fig:lin_comb_grp_100_180})\&(\ref{fig:lin_comb_bad_100_180}) show the linear combination search for $(m,n) = (100,180)$.  The model was retrained, and achieved $87.1\%$ for the displayed results.

At lower dimensions, the optimal latent dimension was no longer shared between the satisfied conditions.

The unsatisfied conditions displayed mostly the same behavior at lower dimensions.

\subsection{Embedding structure}
To explore the geometric distribution of word vectors, the angles and distances between 1) random pairs of words, 2) all words and the global average word vector, 3) random pairs of co-occurring words, 4) all words with a co-occurring word vector average, 5) adjacent tangent vectors, 6) tangent vectors with a co-occurring tangent vector average were computed. The magnitudes of the average word vectors, average co-occurring word vectors, and average tangent vectors were also computed.

Additionally, the relative effect of words on states is computed verses their cosine similarities and relative distances, measured by Eqs.(\ref{eqn:frac_dist})-(\ref{eqn:dist_sim}).

In the figures that follow, there are, generally, three categories of word vectors explored: 1) random word vectors from the pool of all word vectors, 2) co-occurring word vectors, and 3) tangent vectors (the difference vector between adjacent words).

\begin{figure}[H]
\begin{center}
\includegraphics[scale=0.5]{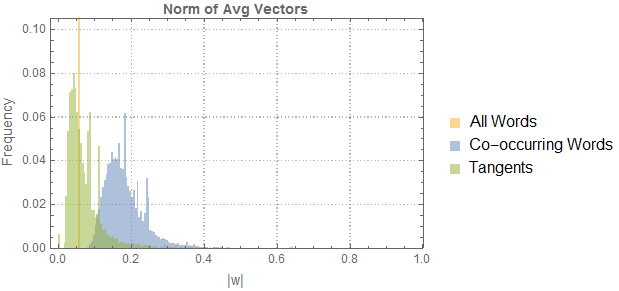}
\caption{\footnotesize The frequency distribution of the norm of average vectors.  There was one instance of a norm for the average of all word vectors, hence the singular spike for its distribution.  The other vector distributions were over the average for different individual tweets.} \label{fig:norm_dist}
\end{center}
\end{figure}

Fig.(\ref{fig:norm_dist}) shows the distribution in the Euclidean norms of the average vectors that were investigated.
The tangent vectors and average word vectors had comparable norms.  The non-zero value of the average word vector indicates that words do not perfectly distribute throughout space.  The non-zero value of the average tangent vectors indicates that tweets in general progress in a preferred direction relative to the origin in embedding space; albeit, since the magnitudes are the smallest of the categories investigated, the preference is only slight.  The norm of the average of co-occurring word vectors is significantly larger than the norms of others categories of vectors, indicating that the words in tweets typically occupy a more strongly preferred region of embedding space (e.g. in a cone, thus preventing component-wise cancellations when computing the average).

\begin{figure}[H]
\begin{center}
\begin{tabular}{cc}
\sidesubfloat[]{\includegraphics[scale=0.5]{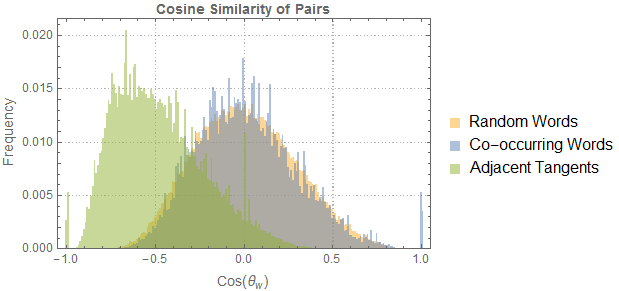}}\\
\sidesubfloat[]{\includegraphics[scale=0.5]{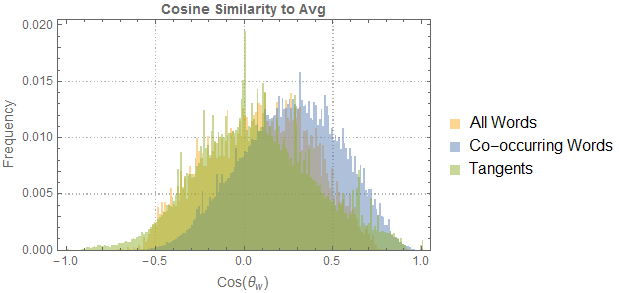}}
\end{tabular}
\caption{\footnotesize Distributions of cosine similarities of vectors with respect to (a) other vectors (b) category average vectors. Averages were taken as they were in Fig.(\ref{fig:norm_dist}). } \label{fig:cos_sim_dist}
\end{center}
\end{figure}

Fig.(\ref{fig:cos_sim_dist}) shows the distribution of the Euclidean cosine similarities of both pairs of vectors and vectors relative to the categorical averages.
The cosine similarity of pairs of random words and co-occurring words shared a very common distribution, albeit with the notable spikes are specific angles and a prominent spike at $\cos(\theta_w) = 1$ for co-occurring pairs.  The prominent spike could potentially be explained by the re-occurrence of punctuation within tweets, so it may not indicate anything of importance; the potential origin of the smaller spikes throughout the co-occurring distribution is unclear.  Generally, the pairs strongly preferred to be orthogonal, which is unsurprising given recent investigations into the efficacy of orthogonal embeddings \cite{Choromanski:orthoembed}.  Adjacent pairs of tangent vectors, however, exhibited a very strong preference for obtuse relative angles, with a spike at $\cos(\theta_w) = -1$.

Words tended to have at most a very slightly positive cosine similarity to the global average, which is again indicative of the fact words did not spread out uniformly. Co-occurring words tended to form acute angles with respect to the co-occurring average. Meanwhile, tangent vectors strongly preferred to be orthogonal to the average.

The strong negative cosine similarity of adjacent tangent vectors, and the strong positive cosine similarity of words with their co-occurring average, indicate co-occurring words \textit{tended} to form a grid structure in a cone. That is, adjacent words tended to be perpendicular to each other in the positive span of some set of word basis vectors.  Of course, this was not strictly adhered to, but the preferred geometry is apparent.

\begin{figure}[H]
\begin{center}
\begin{tabular}{cc}
\sidesubfloat[]{\includegraphics[scale=0.5]{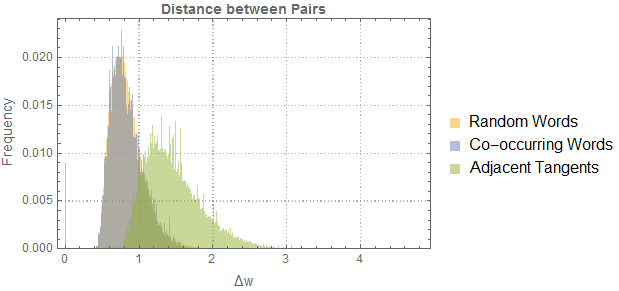}}\\
\sidesubfloat[]{\includegraphics[scale=0.5]{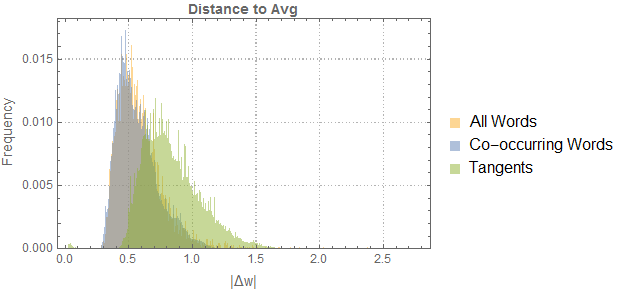}}
\end{tabular}
\caption{\footnotesize Distributions of the Euclidean distances of vectors to (a) other vectors (b) category average vectors. Averages were taken as they were in Fig.(\ref{fig:norm_dist}).} \label{fig:dist_dist}
\end{center}
\end{figure}

Fig.(\ref{fig:dist_dist}) shows the distribution of the Euclidean distances of both pairs of vectors and vectors relative to the categorical averages.
Distributions of random pairs of words and co-occurring words were virtually identical in both plots, indicating that most of the variation is attributable to the relative orientations of the vectors rather than the distances between them.

\begin{figure}[H]
\begin{center}
\begin{tabular}{cc}
\sidesubfloat[]{\includegraphics[scale=0.5]{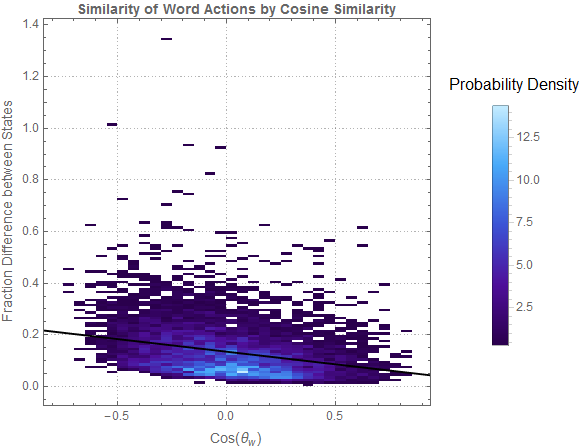}}\\
\sidesubfloat[]{\includegraphics[scale=0.5]{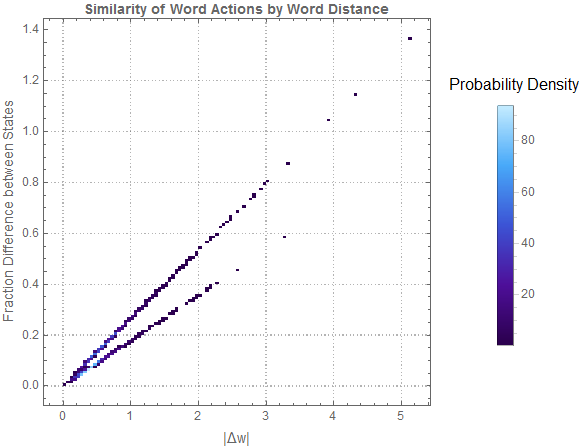}}
\end{tabular}
\caption{\footnotesize Plots of $\mathfrak{E}$ with respect to (a) word cosine similarity, $\cos(\theta_w)$ (b) distance between words, $|\Delta w|$.  Eqs.(\ref{eqn:frac_dist})-(\ref{eqn:dist_sim})} \label{fig:word_sim}
\end{center}
\end{figure}

Fig.(\ref{fig:word_sim}) shows the correlation of the similarity of the action of pairs of words to their cosine similarity and distances apart.
Both plots confirm that the more similar words are, the more similar their actions on the hidden states are.  The strongly linear, bi-modal dependence of the fractional difference on the distance between words indicates that word distance is a stronger predictor of the relative meaning of words than the popular cosine similarity.

\section{Discussion} \label{sec:discussion}
\subsection{Interpretation of results}
The important take-aways from the results are:
\begin{itemize}
\item The GRU trivially learned an identity `word'.

\item The action of the GRU for any individual word admits an inverse for sufficiently large embedding dimension relative to the hidden dimension.

\item The successive action of the GRU for any arbitrary pair of words is not, generally, equivalent to the action of the GRU for any equivalent third `word'.

\item The commutation of successive actions of the GRU for any arbitrary pair of words is not equivalent to the action of the GRU for any equivalent third `word'.

\item The successive action of the GRU for any co-occurring pair of words is equivalent to the action of the GRU for an equivalent third `word' for sufficiently large embedding dimension relative to the hidden dimension.

\item The successive action of the GRU for any series of co-occuring words is equivalent to the action of the GRU for an equivalent `word' for sufficiently large embedding dimension relative to the hidden dimension.

\item The action of the GRU for any series of co-occurring words admits an inverse for sufficiently large embedding dimension relative to the hidden dimension.

\item Any condition satisfied for a sufficiently large embedding dimension relative to the hidden dimension is true for any pair of dimensions given an appropriate linear combination of the outputs of the GRU projected into an appropriate lower dimension (latent dimension).

\item The axiomatic errors for all satisfied conditions for the most performant models are minimized for specific, shared latent dimensions, and increases away from these latent dimensions; the optimal latent dimension is not shared for sufficiently small embedding dimensions.

\item Models with lower test performance tend to optimally satisfy these conditions for lower latent dimensions.

\item Co-occurring word vectors tend to be perpendicular to each other and occupy a cone in embedding space.

\item The difference of the action of two word vectors on a hidden state increases linearly with the distance between the two words, and follows a generally bi-modal trend.
\end{itemize}

Although there are still several outstanding points to consider, we offer an attempt to interpret these results in this section.

Identity, inverse, and closure properties for co-occurring words are satisfied, and in such a way that they are all related under some algebraic structure.  Since closure is not satisfied for arbitrary pairs of words, there are, essentially, two possible explanations for the observed structure:
\begin{enumerate}
\item The union of all sets of co-occurring words is the Cartesian product of multiple Lie groups:
\be
W = G_1 \times G_2 \times \dots, \label{eqn:cartesian_grp}
\ee
where $W$ is the space of words, and $G_i$ is a Lie group.  Since multiplication between groups is not defined, the closure of arbitrary pairs of words is unsatisfied.
\item The GRU's inability to properly close pairs of words it has never encountered together is the result of the generalization problem, and all words consequently embed in a larger Lie group:
\be
G_1 \times G_2 \times \dots \subset G = W. \label{eqn:total_grp}
\ee
\end{enumerate}
In either case, words can be considered elements of a Lie group.  Since Lie groups are also manifolds, the word vector components can be interpreted as coordinates on this Lie group.  Traditionally, Lie groups are practically handled by considering the Lie algebra that generates them, $\mathfrak{g}, G = \exp(\mathfrak{g})$. The components of the Lie vectors in $\mathfrak{g}$ are then typically taken to be the coordinates on the Lie group.  This hints at a connection between $\mathfrak{g}$ and the word vectors, but this connection was not made clear by any of the experiments.  Furthermore, RNNs learn a nonlinear representation of the group on some latent space spanned by the hidden layer.

Since sentences form paths on the embedding group, it's reasonable to attempt to form a more precise interpretation of the action of RNNs.  We begin by considering their explicit action on hidden states as the path is traversed:
\begin{subequations}
\begin{align}
h_{t+1} =& R_{w_t} h_t \implies \label{eqn:rnn_action} \\
(h_{t+1} - h_t) + \gamma_{w_t} h_t =& 0, \label{eqn:rnn_connection} \\
\gamma_{w_t} \equiv& 1 - R_{w_t}.
\end{align}
\end{subequations}
Eq.(\ref{eqn:rnn_connection}) takes the form of a difference equation.  In particular, it looks very similar to the finite form of the differential equation governing the nonlinear parallel transport along a path, $m(t)$, on a principal fibre bundle with base space $M$ and group $G=\exp(\mathfrak{g})$.  If the tangent vector at $m(t)$ is $v(t)$, and the vector being transported at $m(t)$ is $h(t)$ then we have
\begin{align}
\partial_t h(t) + v^\mu(t) \Gamma_\mu[m(t), h(t)] =&0, \label{eqn:parallel_transport}
\end{align}
where $\Gamma$ is the (nonlinear) connection at $m(t)$.  If $v$ were explicitly a function of $m$, Eq.(\ref{eqn:parallel_transport}) would take a more familiar form:
\begin{subequations}
\begin{align}
\partial_t h(t) + \gamma_{m(t)}[h(t)] =& 0, \label{eqn:geodesic_transport} \\
\gamma_{m(t)} \equiv& v^\mu[m(t)] \Gamma_\mu[m(t), h(t)]. \label{eqn:rnn_connection_relation}
\end{align}
\end{subequations}
Given the striking resemblance between Eqs.(\ref{eqn:geodesic_transport})\&(\ref{eqn:rnn_connection}), is it natural to consider either
\begin{enumerate}
\item The word embedding group serving as the base space, $M=G$, so that the path $m(t)$ corresponds explicitly to the sentence path.
\item A word field on the base space, $w: M \to G$, so that there exists a mapping between $m(t)$ and the sentence path.
\end{enumerate}
The second option is more general, but requires both a candidate for $M$ and a compelling way to connect $m(t)$ and $v(t)$.  This is also more challenging, since, generally, parallel transport operators, while taking values in the group, are not closed. If the path were on $G$ itself, closure would be guaranteed, since any parallel transport operator would be an element of the co-occurring subgroup, and closure arises from an equivalence class of paths.

To recapitulate the final interpretations of word embeddings and RNNs in NLP:
\begin{itemize}
\item Words naturally embed as elements in a Lie group, $G$, and end-to-end word vectors may be related to the generating Lie algebra.

\item RNNs learn to parallel transport nonlinear representations of $G$ either on the Lie group itself, or on a principal $G$-bundle.
\end{itemize}

\subsection{Proposal for class of recurrent-like networks}

The geometric derivative along a path parameterized by $t$ is defined as:
\be
D_t = \partial_t + v^\mu(t) \Gamma_\mu,
\ee
where $v(t)$ is the tangent vector at $t$, and $\Gamma$ is the connection.  This implies RNNs learn the solution of the first-order geometric differential equation:
\be
D_t h(t) = 0.
\ee
It is natural, then, to consider neural network solutions to higher-order generalizations:
\be
D^n_t h(t) = 0. \label{eqn:higher_order}
\ee
Networks that solve Eq.(\ref{eqn:higher_order}) are recurrent-like.  Updates to a hidden state will generally depend on states beyond the immediately preceding one; often, this dependence can be captured by evolving on the phase space of the hidden states, rather than on the sequences of the hidden states themselves.  The latter results in a nested RNN structure for the recurrent-like cell, similar to the structure proposed in \cite{Ruben:nested}.

Applications of Eq.(\ref{eqn:higher_order}) are currently being explored.  In particular, if no additional structure exists and RNNs parallel transport states along paths on the word embedding group itself (the first RNN interpretation), geodesics emerge as a natural candidate for sentence paths to lie on.  Thus, sentence generation could potentially be modeled using the geodesic equation and a nonlinear adjoint representation: $n=2$, $h \in \mathfrak{g}$ in Eq.(\ref{eqn:higher_order}).  This geodesic neural network (GeoNN) is the topic of a manuscript presently in preparation.

\subsection{Proposal for new word embeddings}
The embeddings trained end-to-end in this work provided highly performant results.  Unfortunately, training embeddings on end-tasks with longer documents is challenging, and the resulting embeddings are often poor for rare words. However, it would seem constructing pre-trained word embeddings by leveraging the emergent Lie group structure observed herein could provide competitive results without the need for end-to-end training.

Intuitively, it is unsurprising groups appear as a candidate to construct word embeddings.  Evidently, the proximity of words is governed by their actions on hidden states, and groups are often the natural language to describe actions on vectors.  Since groups are generally non-commutative, embedding words in a Lie group can additionally capture their order- and context-dependence.  Lie groups are also generated by Lie algebras, so one group can act on the algebra of another group, and recursively form a hierarchical tower. Such an arrangement can explicitly capture the hierarchical structure language is expected to exhibit. E.g., the group structure in the first interpretation given by Eq.(\ref{eqn:cartesian_grp}),
\be
G = G_1 \times G_2 \times G_3 \times \dots,
\ee
admits, for appropriately selected $G_N$, hierarchical representations of the form
\begin{subequations}
\begin{align}
R_N :& G_N \times \mathfrak{g}_{N-1} \to \mathfrak{g}_{N-1}, \\
R_1 :& G_1 \times H \to H,
\end{align}
\end{subequations}
where $G_N = \exp(\mathfrak{g}_N)$. Such embedding schemes have the potential to generalize current attempts at capturing hierarchy, such as Poincar\'{e} embeddings \cite{Nickel:poincare}. Indeed, hyperbolic geometries, such as the Poincar\'{e} ball, owe their structure to their isometry groups. Indeed, it is well known that the hyperbolic $N+1$ dimensional Minkowski space arises as a representation of $SO(1,N)$ + translation symmetries.

In practice, Lie group embedding schemes would involve representing words as constrained matrices and optimizing the elements, subject to the constraints, according to a loss function constructed from invariants of the matrices, and then applying the matrix log to obtain Lie vectors. A prototypical implementation, dubbed ``LieGr," in which the words are assumed to be in the fundamental representation of the special orthogonal group, $SO(N)$, and are conditioned on losses sensitive to the relative actions of words, is the subject of another manuscript presently in preparation.

\section{Closing remarks} \label{sec:closing}
The results presented herein offer insight into how RNNs and word embeddings naturally tend to structure themselves for text classification.  Beyond elucidating the inner machinations of deep NLP, such results can be used to help construct novel network architectures and embeddings.

There is, however, much immediate followup work worth pursuing.  In particular, the uniqueness of identities, inverses, and multiplicative closure was not addressed in this work, which is critical to better understand the observed emergent algebraic structure.  The cause for the hyperparameter stratification of the error in, and a more complete exploration of, commutative closure remains outstanding.  Additionally, the cause of the breakdown of the common optimal latent dimension for low embedding dimension is unclear, and the bi-model, linear relationship between the action of words on hidden states and the Euclidean distance between end-to-end word embeddings invites much investigation.

As a less critical, but still curious inquiry: is the additive relationship between words, e.g. ``king - man + woman = queen," preserved, or is it replaced by something new?  In light of the Lie group structure words trained on end tasks seem to exhibit, it would not be surprising if a new relationship, such as the Baker-Campbell-Hausdorff formula\footnote{Since the BCH formula is simply an non-commutative correction to the additive formula usually applied for analogies, it may be possible that this relation would already better represent analogies for disparately-related words.}, applied.

\section{Acknowledgements} \label{sec:acknowledgements}
The author would like to thank Robin Tully, Dr. John H. Cantrell, and Mark Laczin for providing useful discussions, of both linguistic and mathematical natures, as the work unfolded.  Robin in particular provided essential feedback throughout the work, and helped explore the potential use of free groups in computational linguistics at the outset.  John furnished many essential conversations that ensured the scientific and mathematical consistency of the experiments, and provided useful insights into the results.  Mark prompted the investigation into potential emergent monoid structures since they appear frequently in state machines.

\appendix
%\section{Algebraic structures} \label{app:algebra}

%\section{Connections and parallel transport} \label{app:geometry}

%\section{Universal approximation theorem and degeneracy} \label{app:uat}

\bibliographystyle{utphys}
\bibliography{biblio}

\end{document}